\documentclass{aci}

\usepackage{txfonts,subfigure}
\def\typeofarticle{Research article}
\def\currentvolume{5}
\def\currentissue{2}
\def\currentyear{2025}

\def\ppages{xxx--xxx.}
\def\DOI{10.3934/aci.2025xxx}
\def\Received{25 June 2025}
\def\Revised{09 December 2025}
\def\Accepted{11 December 2025}
\def\Published{ December 2025}
\usepackage{cite}
\begin{document}
\title{Multilevel neural networks with dual-stage feature fusion for human activity recognition}
\author{%
Abeer FathAllah Brery\affil{1,}\corrauth, Ascensión Gallardo-Antolín\affil{2}, Israel Gonzalez-Carrasco\affil{1} and Mahmoud Fakhry\affil{3,}\corrauth}
         
\shortauthors{the Author(s)}

\address{%
\addr{\affilnum{1}}{Departamento de Informática, Universidad Carlos III de Madrid, Avenida de la Universidad, 30, Leganés, Madrid, 28911, Spain}
\addr{\affilnum{2}}{Departamento de Teoría de la Señal y Comunicaciones, Universidad Carlos III de Madrid, Avenida de la Universidad, 30, Leganés, Madrid, 28911, Spain} 
\addr{\affilnum{3}}{CEIEC, Universidad Francisco de Vitoria, Ctra.M-515 Pozuelo-Majadahonda Km.1, 800, Pozuelo de Alarcón, Madrid, 28223, Spain}}
\corraddr{Email: abeer\_brery@aswu.edu.eg, mahmoud.fakhry@ufv.es.}

\editor{Xuesong Zhai}

\begin{abstract}
Human activity recognition (HAR) refers to the process of identifying human actions and activities using data collected from sensors. Neural networks, such as convolutional neural networks (CNNs), long short-term memory (LSTM) networks, convolutional LSTM, and their hybrid combinations, have demonstrated exceptional performance in various research domains. Developing a multilevel individual or hybrid model for HAR involves strategically integrating multiple networks to capitalize on their complementary strengths. The structural arrangement of these components is a critical factor influencing the overall performance. This study explores a novel framework of a two-level network architecture with dual-stage feature fusion: late fusion, which combines the outputs from the first network level, and intermediate fusion, which integrates the features from both the first and second levels. We evaluated $15$ different network architectures of CNNs, LSTMs, and convolutional LSTMs, incorporating late fusion with and without intermediate fusion, to identify the optimal configuration. Experimental evaluation on two public benchmark datasets demonstrates that architectures incorporating both late and intermediate fusion achieve higher accuracy than those relying on late fusion alone. Moreover, the optimal configuration outperforms baseline models, thereby validating its effectiveness for HAR.

\end{abstract}
\maketitle

\section{Introduction}
Human activity recognition (HAR) plays a crucial role in numerous domains, including healthcare monitoring, security systems, intelligent environments, and surveillance applications, where accurate interpretation of human movements is critical \cite{Dhekane2024TransferLI,w22}. In recent years, deep learning models have significantly advanced the field by eliminating the need for manual feature engineering and achieving high classification performance through the automatic extraction of complex, discriminative features from raw sensor data. Despite the proliferation of various deep learning architectures for HAR, a systematic and comprehensive evaluation of their comparative performance is often lacking \cite{Intro2, Intro3}. Such assessments are crucial for understanding the strengths and limitations of each model and their generalizability, scalability, and applicability to different types of sensor data, activity categories, and deployment environments in the real world. A thorough comparative analysis can guide researchers and practitioners in selecting the most suitable models for specific HAR scenarios, ultimately contributing to the advancement and practical implementation of robust systems.

Among deep learning models, convolutional neural network (CNN) and long short-term memory (LSTM) architectures are the most widely used. Hybrid models are often designed to outperform individual models by offering benefits such as reduced computation time and the capability to leverage data from various sensor positions \cite{Intro5}. The ability of CNN-LSTM hybrid models to capture both spatial and temporal dependencies is typically achieved by using CNN layers for feature extraction from the input data, followed by LSTM layers for sequence modeling. The process of merging features from different layers or modalities, known as feature fusion, has been explored in recent models and is commonly implemented using operations such as addition or concatenation \cite{He2015DeepRL}.

A comprehensive review of deep learning models used in smartphone and wearable sensor-based recognition systems was provided in \cite{Intro5}. These include models such as CNNs, LSTMs, and various hybrid architectures, each of which is discussed in terms of its unique characteristics, strengths, and limitations. In \cite{U1}, a heterogeneous convolution approach divides the kernels in a CNN into two groups: one that recalibrates the other group. Dynamic CNN introduces dynamic kernels with attention that adapt weights \cite{U5}, and deep convolution constructs an ensemble stream employing late fusion \cite{U26}.  In \cite{U29}, the authors introduced multiscale hierarchical CNNs that incorporate adaptive feature fusion and dynamic channel selection based on LSTM. Hybrid CNN-LSTM models, as proposed in \cite{U14,U9}, are designed to capture spatio-temporal dynamics across multiple sensors. These models can identify key feature embeddings by incorporating self-attention mechanisms into their architecture. 

The hierarchical deep LSTM (H-LSTM) model introduced in \cite{C2} uses the characteristics of the time-frequency domain for HAR. A multi-head CNN architecture was proposed in \cite{MheadCNNlstmUCI}, where three parallel CNNs processed data from different sensors. The outputs were concatenated and passed through the LSTM and dense layers of the model. In \cite{C3}, CNN and LSTM models were evaluated for HAR applications. The raw signals were preprocessed using a Butterworth filter, and nine features were extracted per 128-sample window. The study in \cite{U15} proposed a CNN combined with LSTM to extract features and capture temporal dependencies from accelerometer and gyroscope data. Finally, in \cite{Intro7}, a multilevel feature fusion strategy was introduced for multidimensional HAR. This approach employs a multi-head CNN for visual input and a CNN-LSTM combination to extract temporal features from multisensor time-series data. The architecture incorporates three CNN branches with a channel attention module to enhance the representation of the channel and spatial characteristics.

This study investigates two-level network architectures that employ a feature fusion strategy to integrate features from the same or different network levels for HAR. The experiments explore different combinations of CNN, LSTM, and convolutional LSTM (CLSTM) \cite{Shi2015ConvolutionalLN}. CLSTM differs from conventional CNN-LSTM architectures by integrating convolutional operations directly into the recurrent structure, thereby forming a unified spatio-temporal architecture. This study is the first to apply CLSTM in individual and hybrid configurations for HAR applications.

\section{Datasets}
The rapid growth of wearable technology has allowed the development of various HAR datasets; however, challenges in standardization, sharing, and accessibility often limit their reusability and reproducibility \cite{Intro4}. Choosing an appropriate dataset for a given HAR task involves considering multiple factors, such as the number of participants, the variety of activities, sensor modalities, and the recording environment. To evaluate the performance of the different models and ensure broad applicability and benchmark performance, we selected two well-established and widely used datasets: the USC-HAD dataset \cite{USC-HAD} and the UCI-HAR dataset \cite{UCI-HAR}. These datasets encompass daily living activities and offer diverse activity labels and robust sensor configurations that are suitable for evaluating deep models.

\subsection{USC-HAD}
The University of Southern California Human Activity Dataset (USC-HAD) is a resource for research on human activity in the ubiquitous computing community \cite{USC-HAD}. This dataset includes $14$ subjects and $12$ daily activities, with sensor hardware attached to the right front hip of the subjects. Sensor recordings are the most basic and common human activities, including walking, running, jumping, sitting, sleeping, and using an elevator. To capture variations in activity, each subject was asked to perform $5$ trials for each activity on different days at various indoor and outdoor locations. They used the so-called MotionNode to capture activity signals and build a dataset, which is a multimodal sensor that integrates a three-axis accelerometer, three-axis gyroscope, and three-axis magnetometer at a sampling rate of $100$ Hz. The USC-HAD dataset provides a controlled yet diverse representation of real-world activities, making it a valuable benchmark for evaluating HAR model performance.

\subsection{UCI-HAR}
The University of California Irvine (UCI-HAR) collected this dataset from recordings of $30$ subjects performing $6$ activities while carrying a smartphone mounted with embedded inertial sensors \cite{UCI-HAR}. Each participant was instructed to follow an activity protocol while wearing a Samsung Galaxy S II smartphone mounted on their waists. The six selected activities were standing, sitting, lying down, walking, walking downstairs, and walking upstairs. They collected triaxial linear acceleration and angular velocity signals using a phone accelerometer and gyroscope at a sampling rate of $50$ Hz. The time signals were sampled in sliding windows with a fixed width of $2.56$ s and $50\%$ overlap. A feature vector consists of the mean, correlation, signal magnitude area, autoregression coefficients, energy of different frequency bands, frequency skewness, and the angles between vectors. A total of $561$ features were extracted to describe each activity. Due to its structured design and rich feature set, the UCI-HAR dataset has become a widely used benchmark for evaluating the effectiveness of machine learning and deep learning models in HAR research.

\section{Proposed methodology}
\begin{figure}[H]
	\centering
	\includegraphics[scale=0.6,trim={0cm 0cm 0cm 0cm}]{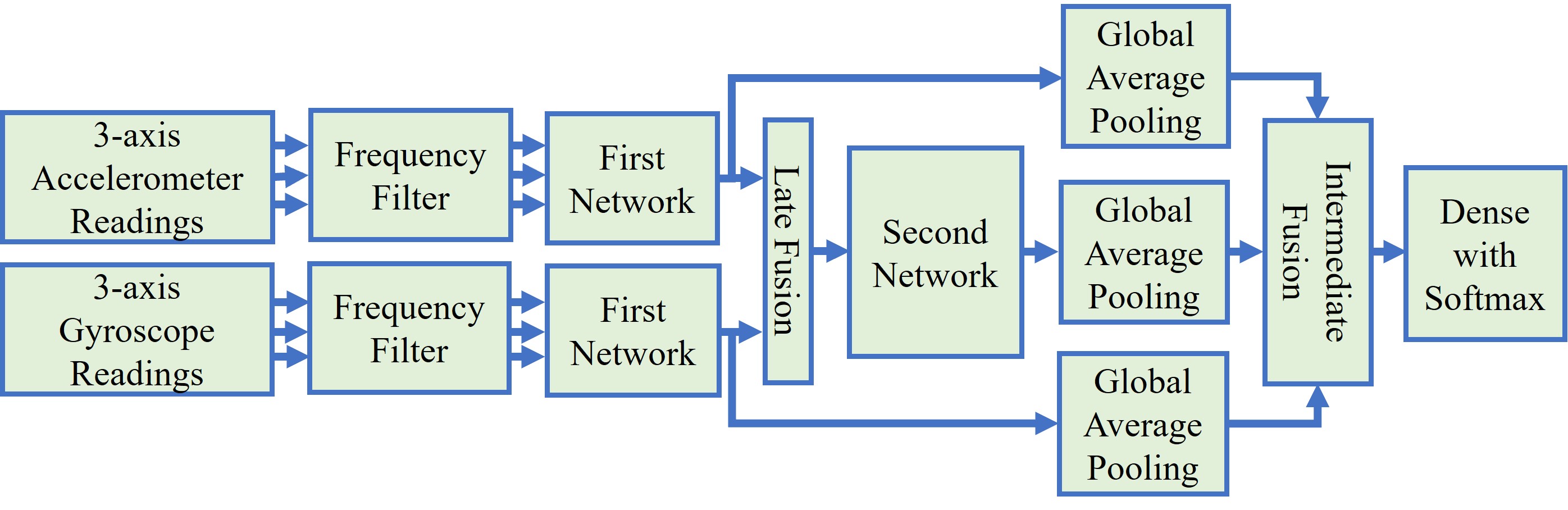}
	\caption{Block diagram of the proposed network architecture.}
	\label{fig:system}
\end{figure}

The block diagram presented in Figure \ref{fig:system} depicts the architecture of the proposed model, which incorporates a feature fusion approach and preprocessing frequency filter. Initially, raw sensor data undergo filtering and normalization to ensure that the model processes low-frequency signals with a zero mean and unit norm. The normalized accelerometer signals, which consisted of three channels, were fed into the first neural network. In parallel, the three-channel normalized gyroscope signals follow a separate but identical processing path to feed another network. The outputs from these two networks are then merged through concatenation to generate either a one- or two-dimensional multichannel feature map. This fusion strategy enables the model to learn high-level representations by integrating complementary information from both accelerometer and gyroscope data.

Subsequently, the feature maps are processed by a second neural network, which introduces an additional abstraction layer into the learned features. To further refine the extracted features, global average pooling is applied to the outputs of both the first and second networks. This operation produces compact yet informative representations by summarizing spatial information. The resulting pooled features are then combined in a concatenation layer to form a unified one-dimensional feature vector that integrates information from both network stages. Finally, the feature vector is processed by a dense layer, followed by a classification softmax layer, to enable effective and nuanced activity recognition.

To optimize the architecture selection, we systematically evaluated multiple configurations for both the first and second networks, selected from the CNN, LSTM, and convolutional LSTM architectures, each implemented in one- and two-dimensional forms. In addition, we investigated the impact of internal feature fusion strategies by comparing models that employed single- or dual-stage fusion mechanisms. This comprehensive evaluation was designed to identify the most effective neural network configuration for optimizing feature integration and improving the ability of the model to accurately recognize HAR based on raw sensor data.

Deep neural networks have emerged as powerful models for learning representations from complex data. Each architecture exhibits unique characteristics tailored to specific tasks, from foundational feedforward networks to more sophisticated CNNs and RNNs, such as LSTM. CNNs excel at extracting spatial hierarchies from images, whereas LSTMs capture the temporal dependencies in sequential data.  In addition, novel architectures have pushed the boundaries of representation learning to structured and relational data. The design of each architecture, including the layer types, activation functions, and connectivity patterns, profoundly affects its expressivity and computational efficiency. 

\subsection{Convolutional neural networks (CNNs)}
Despite challenges such as limited data on group activities, high computational resource demands, data privacy concerns, and edge computing limitations, CNN-based models remain suitable for accurate and efficient HAR system applications \cite{Intro1}. CNNs can learn highly abstracted object features and are suitable for image analysis and recognition \cite{Ghosh2019FundamentalCO}. However, the CNN model also has a layer that can learn the features of sequential data with multiple variables. A typical CNN model comprises a convolutional layer followed by a smoothing rectified linear unit (ReLU), pooling, and batch normalization layers. The convolutional layer is the main component of the CNN, which operates on the principle of sliding windows to reduce computational complexity. In this layer, a kernel filter is used to extract features from the input data. For a given 2D single-channel input matrix \( Y \) and 2D convolutional filter $H$, the output of the convolution operation at position $(i, j)$ is   
\begin{equation}
Z(i, j) = \sum_{m=0}^{H_f-1} \sum_{n=0}^{W_f-1} Y(i+m, j+n) H(m, n) + b,    
\end{equation}
where \( Y(i, j) \) denotes the input feature map,  \( H(m, n) \) is the convolutional filter, and \( b \) is the bias. For an input \(Y \) of size \( H_{\text{in}} \times W_{\text{in}} \times C_{\text{in}} \) and a filter \(H \) of size \( H_{\text{f}} \times W_{\text{f}} \times C_{\text{f}} \) with stride \( S_H \times S_W\) and padding \( P \), the output \(Z \) is of size given by 
\begin{equation}
H_{\text{out}} = \frac{H_{\text{in}} - H_{\text{f}} + 2P}{S_H} + 1, W_{\text{out}} = \frac{W_{\text{in}} - W_{\text{f}} + 2P}{S_W} + 1, ~and~ C_{\text{out}}=C_{\text{f}}.
\end{equation}

The next layer is the pooling layer, which is designed to reduce the size of the feature map and extract dominant features for efficient model training. Several types of pooling operations exist, including max-pooling and average pooling. The batch normalization layers apply a transformation that maintains the output mean close to $0$ and a standard deviation close to $1$. 

\subsection{Long short-term memory (LSTM) network}
LSTM is a special type of RNN that was developed to overcome the weakness of RNN, which cannot learn long-term dependence \cite{Hochreiter1997LongSM}. LSTM consists of memory blocks called cells, which have two states: cell and hidden states. Cells in LSTM are used to make decisions by storing or ignoring information regarding the forget, input, and output gates of the model. The LSTM operates in three stages: In the first stage, the network works with the forget gate to determine the information that must be ignored or stored in the cell states. The calculation starts by considering the input at the current time step $X_t$ and the previous value of the hidden state $H_{t-1}$ using the sigmoid function $\sigma$, such as
\begin{equation}
f_t = \sigma (W_f [X_t,H_{t-1}]+b_f),
\end{equation}
where $W_f$ and $b_f$ denote the weight and bias of the forget gate, respectively. In the second phase, the network converts the old cell state $C_{t-1}$ into the new cell state $C_t$. This process selects new information in the long-term memory (cell state). To obtain the new cell state value, the calculation process considers the reference values from the forget, input, and cell update gates, as follows: 
\begin{equation}
i_t = \sigma (W_i [X_t,H_{t-1}]+b_i)
\end{equation}
\begin{equation}
C_t=f_t\circ C_{t-1}+i_t\circ \text{tanh}(W_c [X_t,H_{t-1}]+b_c)
\end{equation}
where $i_t$, $W_i$, and $b_i$ denote the output, weight, and bias of the input gate, respectively. $W_c$ and $b_c$ denote the weights and biases of the cell states, respectively. The symbol $\circ$ denotes a point product operation. Once the cell status update is completed, the final step is to determine the value of the hidden state $H_t$. This process aims for the hidden state to act as a memory, containing information about previous data, and to be used for making predictions. To determine the value of the hidden state, the calculation must have the reference value of the new cell state $C_t$ and the output gate $o_t$ in terms of the weights $W_o$ and the bias $b_o$ of the output gate, such as
\begin{equation}
o_t = \sigma (W_o [X_t,H_{t-1}]+b_o),
\end{equation}
\begin{equation}
H_t = o_t \circ \text{tanh}(C_t).
\end{equation}
\subsection{Convolutional LSTM (CLSTM)}
\begin{figure}[H]
	\centering
	\includegraphics[scale=1,trim={0cm 0cm 0cm 0cm}]{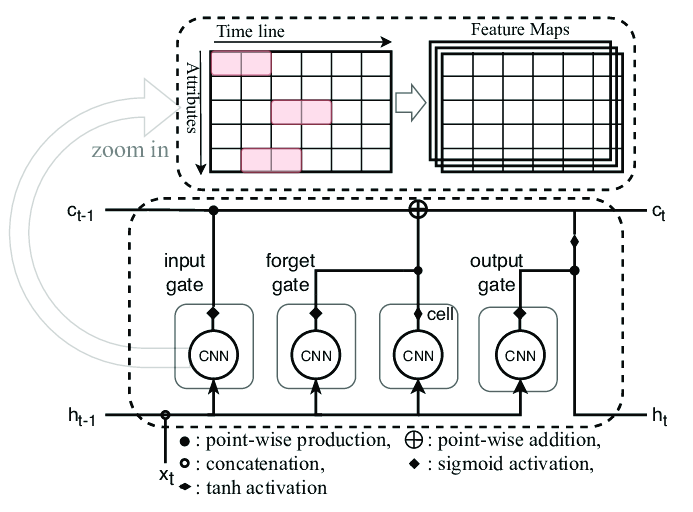}
	\caption{Block diagram of convolutional LSTM network.}
	\label{fig:convlstm}
\end{figure}

Building upon the recurrent framework of LSTM, convolutional LSTM (CLSTM) distinguishes itself by employing convolution operations instead of internal matrix multiplications \cite{Shi2015ConvolutionalLN}. This architectural shift enables the CLSTM to process data while preserving its spatial dimensions (as illustrated in Figure \ref{fig:convlstm}), avoiding the reduction to a flat feature vector that is characteristic of the standard LSTM. This spatial retention is particularly advantageous for tasks involving grid-like data, such as images and video frames. Formally,
\begin{align}
& f_t = \sigma (W_f*[X_t,H_{t-1}]+b_f),\\
& i_t = \sigma (W_i*[X_t,H_{t-1}]+b_i),\\
& C_t=f_t\circ C_{t-1}+i_t\circ \text{tanh}(W_c*[X_t,H_{t-1}]+b_c),\\
& o_t = \sigma (W_o*[X_t,H_{t-1}]+b_o),\\
&H_t = o_t \circ \text{tanh}(C_t),
\end{align}
where * denotes the convolution operator. Integrating convolutional operations within the memory cell and gate computations in a neural network is a significant advancement in its ability to autonomously capture spatial hierarchies and patterns in the input data. This architectural enhancement enables the network to learn local patterns and correlations within the data, thereby fostering a nuanced understanding of spatial context. Convolutional operations within a memory cell facilitate the extraction of spatial features at different scales, thereby allowing the model to discern both fine- and coarse-grained spatial hierarchies. Additionally, the use of convolutional structures in gate computations helps the model selectively focus on relevant spatial information, promoting more effective and context-aware learning.
\subsection{Global average pooling}
Global Average Pooling (GAP) is a pooling operation used in convolutional neural networks (CNNs) that reduces the spatial dimensions of feature maps to a single value per channel. Unlike traditional pooling methods, such as max or average pooling, which partially reduce dimensionality, global average pooling collapses each feature map into a single number by taking the average of all elements in the feature map. Suppose we have a feature map tensor of shape $(H_{in}, W_{in}, C)$, where $ H_{in} $ is the height of the input feature map, $ W_{in} $ is the width of the input feature map, and $C$ is the number of channels. For each channel $ c \in \{1, 2, ..., C\} $, the GAP computes the average value across the spatial dimensions $ H_{in} \times W_{in} $. The output for each channel is given by
\begin{equation}
y_c = \frac{1}{H_{in} \cdot W_{in}} \sum_{i=1}^{H_{in}} \sum_{j=1}^{W_{in}} x(i, j, c),
\end{equation}
where $ x(i, j, c) $ is the value at the position $(i, j)$ in channel $c$, $y_c$ is the resulting average scalar for channel $c$. After applying the GAP, the spatial dimensions are reduced to $1\times1$, and the output tensor has the shape $(1, 1, C)$, which can be interpreted as a vector of size $C$. This operation typically aggregates spatial information to reduce the number of model parameters and prevent overfitting. 

\section{Experimental setup}
The primary objective of this experimental analysis is to conduct a comprehensive ablation study to identify the optimal configurations for the initial and secondary networks. This evaluation is performed under late feature fusion settings, with and without the incorporation of intermediate feature fusion. The investigation involves exhaustively evaluating various architectural combinations, including CNN, LSTM, and CLSTM, while effectively ablating two key design choices: core architectural components and fusion strategies. Concurrently, training hyperparameters must be carefully tuned, as both architectural and parametric choices substantially influence classification performance. To provide a robust evaluation, the performance is assessed using metrics such as accuracy, enabling a comprehensive understanding of how different configurations behave under various conditions and datasets. Finally, the optimal model derived from this ablation study is benchmarked against state-of-the-art methods to validate its effectiveness.

\subsection{Network implementation}
Table \ref{tab:net} lists the key parameters governing the configuration of each network, providing essential insights into the critical choices required for effective optimization. These parameters directly influence the training dynamics and final model performance. Specifically, the table lists the optimization algorithm (ADAM), which controls how weights are updated during learning; the loss function (categorical cross-entropy), which is essential for quantifying the error between the predicted and true labels in classification tasks; and the batch size, which determines the number of samples processed before each internal model update. Understanding these settings is fundamental for reproducing and interpreting network behavior.
\begin{table}[H]
\centering
\caption{Network implementation.}
\label{tab:net}
\begin{tabular}{ll}
\hline
\textbf{Parameter}&\textbf{Value} \\
\hline
\text{Raw sensor readings for USC-HAD} & $2\quad 1024 \times 3$\\
\text{Raw sensor readings for UCI-HAR} & $2\quad 128 \times 3$\\
\text{Input feature vector for UCI-HAR} &$1\quad  561 \times 1$\\
\textbf{1D CNN \& 1D CLSTM:} &\\
\text{no. filters}    &$128$\\
\text{Filter length}  &$16$\\
\text{Filter stride}  &$8$\\
\text{Activation}     &\text{RelU} \\
\textbf{2D CNN \& 2D CLSTM:} &\\
\text{no. filters}    &$128$\\
\text{Filter length}  &$2\times8$\\
\text{Filter stride}  &$2\times4$\\
\text{Activation}     &\text{RelU} \\
\textbf{LSTM:} &\\
\text{no. units}   &$128$\\
\textbf{Training:} &\\
\text{Optimizer}   &\text{ADAM}\\
\text{Loss}        &\text{categorical crossentropy}\\
\text{Batch size}  &no. training examples/32\\
\text{no. epochs}  &$500$\\
\hline
\end{tabular}
\end{table}

\subsection{Evaluation metrics}
Two types of errors can arise: false negatives (FN), when activities belonging to a specific class are incorrectly classified as belonging to other classes, and false positives (FP), when activities from other classes are incorrectly identified as belonging to a specific class. True positives (TP) are activities correctly identified as belonging to the intended class, whereas true negatives (TN) are activities correctly classified as not belonging to that class. These values are crucial for calculating various classification metrics such as the accuracy \cite{Opitz2024ACL}, i.e.,
\begin{equation}
\textbf{Accuracy} = \frac{TP+TN}{TP+TN+FP+FN}\%.
\end{equation}

\section{Experiments and results}
\begin{figure}[H]
	\centering
	\includegraphics[scale=0.7,trim={0cm 0cm 0cm 0cm}]{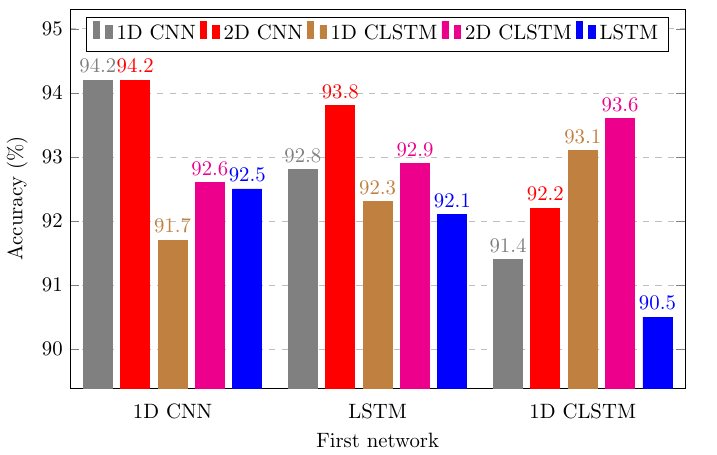}
	\caption{Accuracy of different architectures on the raw sensor readings of the dataset USC-HAD with late feature fusion.}
	\label{fig:ACC11}
\end{figure}

\begin{figure}[H]
	\centering
	\includegraphics[scale=0.7,trim={0cm 0cm 0cm 0cm}]{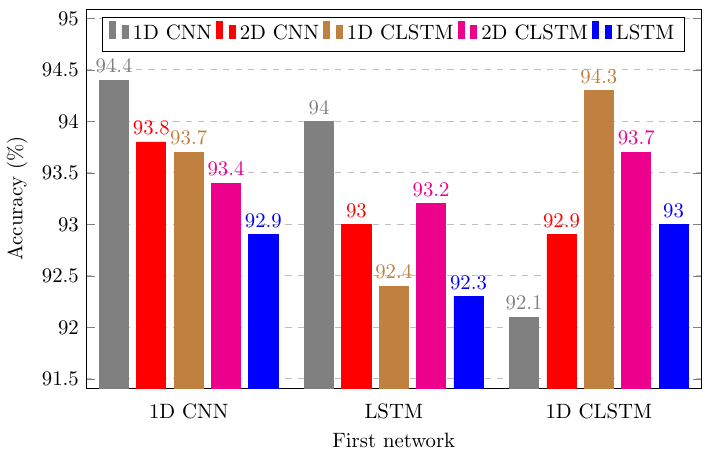}
	\caption{Accuracy of different architectures on the raw sensor readings of the dataset USC-HAD with late and intermediate feature fusion.}
	\label{fig:ACC12}
\end{figure}

We used a $5$-fold data split, a widely adopted approach that balances computational efficiency with reliable performance estimates. The USC-HAD dataset was partitioned into five subsets, and the model was trained and evaluated five times, each time using a different subset as the test set and the remaining four as the training sets. With $14$ subjects and $12$ activities, this splitting produced $672$ ($80\%$) data recordings for model training and $168$ ($20\%$) for testing. Each data recording comprised three vectors representing the measurements from the three-axis accelerometer and three vectors from the three-axis gyroscopes. The vector length is standardized to $1024$ samples, which is achieved by truncating long vectors or replicating samples for shorter vectors. Figures \ref{fig:ACC11} and \ref{fig:ACC12} depict the average accuracy achieved by different network combinations evaluated with and without the intermediate feature fusion strategy.

\begin{figure}[H]
	\centering
	\includegraphics[scale=0.7,trim={0cm 0cm 0cm 0cm}]{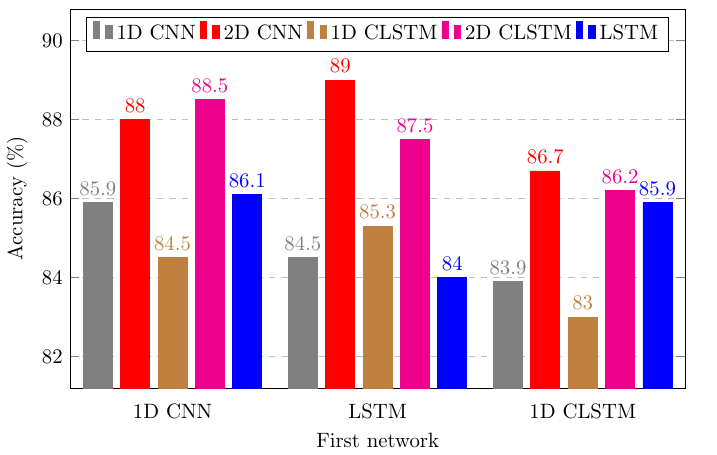}
	\caption{Accuracy of different architectures on the raw sensor readings of the dataset UCI-HAR with late feature fusion.}
	\label{fig:ACC21}
\end{figure}

\begin{figure}[H]
	\centering
	\includegraphics[scale=0.7,trim={0cm 0cm 0cm 0cm}]{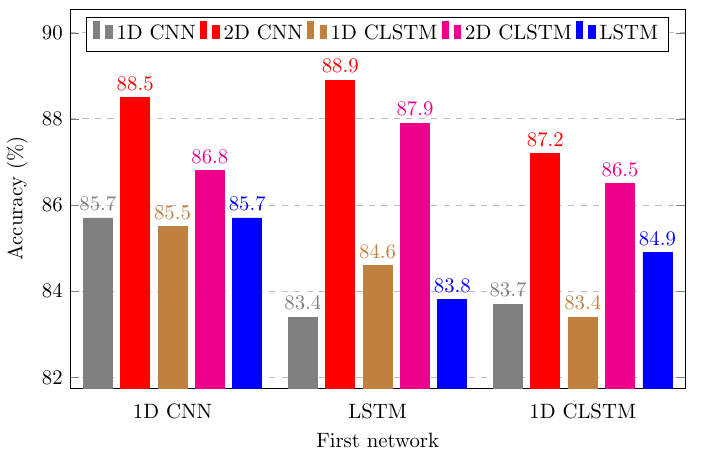}
	\caption{Accuracy of different architectures on the raw sensor readings of the dataset UCI-HAR with late and intermediate feature fusion.}
	\label{fig:ACC22}
\end{figure}

We conducted experiments using either raw sensor readings or commonly used feature vectors from the UCI-HAR dataset to characterize individual activities. The dataset, which contained six distinct activity classes, was partitioned into training and testing sets, resulting in $7352$ data samples (either raw sensor readings or feature vectors) used for training the model and $2947$ samples reserved for testing. Figures \ref{fig:ACC21} and \ref{fig:ACC22} present the test accuracy obtained using raw sensor readings across various combinations of network architectures evaluated using the two fusion strategies. They provide a visual comparison of how the integration of information at different processing stages affects overall classification performance. Table \ref{tab:resUCI} summarizes the detailed accuracy results for each network combination on commonly used feature vectors, allowing for a more granular analysis of the configurations that yield the best recognition rates for the UCI-HAR dataset.


\begin{table}[H]
	\centering
	\caption{Accuracy ($\%$) on features of the dataset UCI-HAR.}
	\label{tab:resUCI}
	\begin{tabular}{llll}
		\hline
		\textbf{First Net}&\textbf{Second Net}&\textbf{No intermediate}&\textbf{With intermediate}\\
		\hline
		\text{1D CNN}  &\text{1D CNN}  &96.30&96.75\\
		\text{1D CNN}  &\text{LSTM}    &94.52&94.65\\
		\text{1D CNN}  &\text{1D CLSTM}&95.76&95.81\\
		\text{LSTM}    &\text{LSTM}    &89.40&87.36\\
		\text{LSTM}    &\text{1D CNN}  &92.45&92.00\\
		\text{LSTM}    &\text{1D CLSTM}&93.80&92.90\\
		\text{1D CLSTM}&\text{1D CLSTM}&95.75&95.54\\
		\text{1D CLSTM}&\text{1D CNN}  &95.33&95.15\\
		\text{1D CLSTM}&\text{LSTM}    &93.61&92.91\\
		\hline
	\end{tabular}
\end{table}
\subsection{Discussion}
Figures \ref{fig:ACC11} and \ref{fig:ACC12} show the accuracy of various architectures on the raw sensor readings of the USC-HAD dataset, highlighting the effectiveness of two-stage individual or hybrid networks with or without intermediate feature fusion. In general, fusion improves the accuracy for most network stage pairings; however, some combinations show only marginal improvements or even slight decreases in accuracy when fusion is applied. The highest accuracy of $94.40\%$ is attained when both the first and second network stages are composed of 1D CNNs with intermediate fusion. The second and third-highest accuracies are achieved by architectures using individual 1D CLSTMs or a hybrid combination of LSTM and 1D CNN, both with fusion. Comparable accuracy is observed for models based on individual 1D CNNs, as well as hybrid combinations of 1D and 2D CNNs or LSTM with 2D CNN with no fusion.

Notably, architectures based on 1D operations (1D CNN, 1D CLSTM) consistently outperformed their 2D counterparts. This indicates that for the inertial measurement unit (IMU) data used in this study, which is fundamentally a one-dimensional temporal signal, 1D convolutions are more effective at extracting discriminative features. Although 2D architectures can learn intersensor correlations, they introduce additional complexity without a commensurate performance gain. This trade-off underscores the advantages of 1D models, which provide superior efficiency and accuracy for the target task.

Figures \ref{fig:ACC21} and  \ref{fig:ACC22} show the accuracy of various models on the raw sensor readings of the UCI-HAR dataset, comparing the models with and without intermediate fusion. The results indicate that network architectures without fusion achieve accuracies ranging from $83\%$ to $89\%$, with the highest performance observed for a hybrid combination of LSTM and 2D CNN. Models that incorporate intermediate fusion generally exhibit enhanced accuracy compared to their non-fused counterparts, underscoring the advantages of combining complementary features extracted by different networks. The top-performing models achieve accuracies of $88.90\%$ and $88.50\%$, realized through the integration of LSTM with 2D CNN and 1D CNN with 2D CNN, respectively, using intermediate fusion.

Table \ref{tab:resUCI} presents the accuracies of the different network architectures on the feature vectors of the UCI-HAR dataset. The highest overall accuracy of $96.75\%$ is achieved when using a 1D CNN, followed by another 1D CNN with the intermediate layer enabled. In general, configurations involving~1D CNNs tend to outperform those involving LSTM or 1D CLSTM as the second stage, suggesting that the convolutional layers are more effective in capturing spatial features in this context. Additionally, the use of intermediate fusion slightly improves the accuracy across most configurations, except in cases involving LSTM as the second network, where it often leads to a reduced improvement. These results highlight the importance of network architecture and feature fusion in achieving optimal accuracy.

\subsection{Comparison with state-of-the-art methods}
To benchmark the performance of the proposed dual-stage fusion architecture, we compared it with several established and recent state-of-the-art (SOTA) methods from the literature on the two benchmark datasets. This comparison is crucial for validating the effectiveness and competitiveness of the proposed approach. Table 3 presents the classification accuracies of the different deep models.

The table includes performance results from a heterogeneous CNN (CNN+HC) that uses grouped kernels for recalibration \cite{U1}, a hierarchical LSTM (H-LSTM) designed for time-frequency characteristics \cite{C2}, and standard CNN-LSTM models that combine spatial and temporal feature extractors  \cite{U9,C3,U14}. Furthermore, we compare against a more complex multi-head CNN-LSTM variant \cite{MheadCNNlstmUCI}, which uses parallel CNNs for different sensors before temporal modeling, representing a strong and sophisticated baseline.

On the USC-HAD dataset, our proposed approach (a dual-stage 1D CNN with intermediate fusion) achieves the highest reported accuracy of $94.40\%$, outperforming all existing methods, including the complex CNN+HC and various CNN-LSTM implementations. Similarly, for the UCI-HAR dataset, our model reaches an impressive accuracy of $96.75\%$, surpassing not only H-LSTM and standard CNN-LSTM architectures but also the more advanced multi-head CNN-LSTM variant.

Notably, many of the compared methods only report results on one of the two datasets, whereas our model demonstrates consistent and superior performance across both. This underscores its strong generalization capability and robustness, validating the effectiveness of the proposed dual-stage feature fusion architecture against a range of SOTA benchmarks.
\begin{table}[H]
\centering
\caption{Accuracy ($\%$) comparison.}
\begin{tabular}{lll}
\hline
\textbf{Method}&\textbf{USC-HAD}&\textbf{UCI-HAR} \\
\hline
\text{CNN+HC\cite{U1}}&90.67&\text{-} \\
\text{H-LSTM\cite{C2}}&\text{-}&91.65 \\
\text{CNN-LSTM}&\text{\cite{U9} 90.88}&\text{\cite{C3} 92.83} \\
\text{}&\text{\cite{U14} 90.91}&\text{ \cite{MheadCNNlstmUCI} 95.76} \\
\textbf{Proposed}&\textbf{94.40}&\textbf{96.75} \\
\hline
\end{tabular}
\label{tab:com}
\end{table}
\section{Conclusions and future work}
This study systematically explored a human activity recognition (HAR) system that employs late and intermediate feature fusion within individual and hybrid deep neural network models. Through a comprehensive ablation study, we evaluated the impact of architectural components (CNN, LSTM, CLSTM) and fusion strategies across $15$ different network configurations. By integrating multimodal data from multiple sensors and combining features across different network levels, the proposed methodology investigates the structures of various architectures. The experimental results on the two benchmark HAR datasets demonstrated the superiority of convolutional modeling using CNNs and convolutional LSTMs over linear modeling using LSTMs. Moreover, our comprehensive evaluation revealed that 1D architectural components (1D CNN and 1D CLSTM) were consistently more effective than 2D components for processing the inherent temporal structure of IMU sensor data, achieving the highest accuracy with greater parameter efficiency. Furthermore, the ablation study conclusively demonstrated that fusing features at an intermediate stage consistently enhanced the classification accuracy over using late fusion alone. Finally, benchmarking against state-of-the-art methods confirmed that our optimal model (dual 1D CNNs with intermediate fusion) achieved superior performance on both datasets, highlighting its effectiveness and generalizability.

This study established a strong baseline using concatenation for its simplicity and effectiveness in our architectural investigation. Future research will explore more sophisticated, adaptive feature fusion mechanisms, such as attention-based fusion, to dynamically weight the contributions from different sensors and network levels, potentially further boosting performance and robustness. Furthermore, although the proposed dual-stage architecture demonstrates high accuracy, its computational complexity presents a challenge for deployment on low-power edge computing devices. Therefore, future work will focus on optimizing this framework using techniques such as model pruning, quantization, and neural architecture search to reduce its memory footprint and latency, facilitating its application in real-time HAR systems.

\section*{Use of AI tools declaration}
The authors declare that they have not used Artificial Intelligence (AI) tools in the creation of this article.

\section*{Conflict of interest}
The authors declare there is no conflicts of interest.

\end{document}